\documentclass[11pt,a4paper]{article}
\usepackage[hyperref]{acl2020}
\usepackage{times}
\usepackage{latexsym}
\usepackage{paralist}

\usepackage{microtype}

\usepackage{latexsym}
\usepackage{adjustbox}

\usepackage{afterpage}
\usepackage[section]{placeins}

\usepackage{amsmath,amsfonts,amssymb }
 
\usepackage{relsize}
\usepackage{multirow}
\usepackage{url}
\usepackage{pgf}
 
\usepackage{latexsym}
\usepackage{graphicx} 
\usepackage{siunitx} 
\usepackage{booktabs} 
\usepackage{xspace}  
\usepackage{soul}
\usepackage{caption} 
\usepackage{rotating} 
\usepackage{placeins}  
\usepackage{subcaption}

\usepackage{url}
\usepackage{enumitem}
\usepackage[T1]{fontenc} 
\usepackage[utf8x]{inputenc} 
\aclfinalcopy 
\def\aclpaperid{3430} 

\newcommand{\secref}[2][]{Section#1~\ref{#2}\xspace}

\newcommand{\figref}[2][]{Figure#1~\ref{#2}\xspace}

\newcommand{\metric}[1]{\textsc{#1}\xspace}
\newcommand{\bleu}{\metric{BLEU}}

\newcommand{\esim}{\metric{ESIM}}
\newcommand{\yisi}{\metric{YiSi-1}}
\newcommand{\yisiqe}{\metric{YiSi-2}}

\newcommand{\ter}{\metric{TER}}

\newcommand{\chrf}{\metric{chrF}}

\title{Tangled up in \textsc{Bleu}: Reevaluating the Evaluation of Automatic Machine Translation Evaluation Metrics}

\author{Nitika Mathur \qquad Timothy Baldwin \qquad Trevor Cohn\\
School of Computing and Information Systems\\
  The University of Melbourne\\
  Victoria 3010, Australia\\[1ex]
  \texttt{nmathur@student.unimelb.edu.au}\quad \texttt{{\{tbaldwin,tcohn\}}@unimelb.edu.au}
  }
 
\date{}

\begin{document}
\maketitle

\begin{abstract}
Automatic metrics are fundamental for the development and evaluation of machine translation systems.
Judging whether, and to what extent, automatic metrics concur with the gold standard of human evaluation is not a straightforward problem.
We show that current methods for judging metrics are highly sensitive to the translations used for assessment, particularly the presence of outliers, which often leads to falsely confident conclusions about a metric's efficacy.  
Finally, we turn to pairwise system ranking, developing a method for thresholding performance improvement under an automatic metric against human judgements, which allows quantification of type I versus type II errors incurred, i.e., insignificant human differences in system quality that are accepted, and significant human differences that are rejected.
Together, these findings suggest improvements to the protocols  for metric evaluation and system performance evaluation in machine translation.
 
\end{abstract} 

\section{Introduction}

Automatic metrics are an indispensable part of machine translation (MT) evaluation, serving as a proxy to human evaluation which is considerably more expensive and time-consuming. They provide immediate feedback during MT system development 
and serve as the primary metric to report the quality of MT systems.  
Accordingly, the reliability of metrics is critical to progress in MT research.

A particularly worrying finding was made in the most recent Conference on Machine Translation (WMT),
as part of their annual competition findings to benchmark progress in translation and translation evaluation.
WMT has established a method based on Pearson's correlation coefficient for measuring how well automatic metrics match with human judgements of translation quality, which is used to rank metrics and to justify their widespread use in lieu of human evaluation.
Their findings \cite{WMT19metrics} showed that if the correlation is computed for metrics using a large cohort of translation systems, typically very high correlations were found between leading metrics and humans (as high as $r = 0.9$). However, if considering only the few best systems, the correlation reduced markedly. This is in contrast to findings at sentence-level evaluation, where metrics are better at distinguishing between high-quality translations compared to low-quality translations \cite{fomicheva2019taking}.

When considering only the four best systems, the automatic metrics were shown to exhibit negative correlations in some instances. 
It would appear that metrics can only be relied upon for making coarse distinctions between poor and good translation outputs, but not for assessing similar quality outputs, i.e., the most common application faced when assessing incremental empirical improvements.

Overall these findings raise important questions as to the reliability of the accepted best-practises for ranking metrics, and more fundamentally, cast doubt over these metrics' utility for tuning high-quality systems, and making architecture choices or publication decisions for empirical research.

In this paper, we take a closer look into this problem, using the metrics data from recent years of WMT to answer the following questions:
\begin{enumerate}
\item Are the above problems identified with Pearson's correlation evident in other settings besides small collections of strong MT systems? To test this we consider a range of system quality levels, including random samples of systems, and show that the problem is widely apparent.  
\item What is the effect of outlier systems in the reported correlations? Systems that are considerably worse than all others can have a disproportionate effect on the computed correlation, despite offering very little insight into the evaluation problem. We identify a robust method for identifying outliers, and demonstrate their effect on correlation, which for some metrics can result in radically different conclusions about their utility.
\item Given these questions about metrics' utility, can they be relied upon for comparing two systems? More concretely, we seek to quantify the extent of improvement required under an automatic metric such that the ranking reliably reflects human assessment. In doing so, we consider both type I and II errors, which correspond to accepting negative or insignificant differences as judged by humans, versus rejecting human significant differences; both types of errors have the potential to stunt progress in the field.  
\end{enumerate}

Overall we find that current metric evaluation methodology can lend false confidence to the utility of a metric, and that leading metrics require either untenably large improvements to serve a gatekeeping role, or overly permissive usage to ensure good ideas are not rejected out of hand. Perhaps unsurprisingly, we conclude that metrics are inadequate as a substitute for human evaluations in MT research.~\footnote{Code, data and additional analysis available at https://github.com/nitikam/tangled}


\section{Related work}
   
Since 2007, the Conference on Machine Translation (WMT) has organized an annual shared task on  automatic metrics, where metrics are evaluated based on correlation with human judgements over a range of MT systems that were submitted to the translation task. Methods for both human evaluation and meta evaluation of metrics have evolved over the years.

In early iterations, the official evaluation measure was the Spearman's rank correlation of metric scores with human scores \cite{callison2006reevaluating}. However, many MT system pairs have very small score differences, and evaluating with Spearman's correlation harshly penalises metrics that have a different ordering for these systems. This was replaced by the Pearson correlation in 2014 \cite{wmt14}. To test whether the difference in the performance of two metrics is statistically significant, the William's test for dependent correlations is used \citep{graham2014testing}, which takes into account the correlation between the two metrics. Metrics that are not outperformed by any other metric are declared as the winners for that language pair. 
  
Pearson's $r$ is highly sensitive to outliers \cite{osborne2004power}: even a single outlier can have a drastic impact on the value of the correlation coefficient; and in the extreme case, outliers can give the illusion of a strong correlation when there is none, or mask the presence of a true relationship. More generally, very different underlying relationships between the two variables can have the same value of the correlation coefficient \cite{anscombe1973graphs}.\footnote{\url{https://janhove.github.io/teaching/2016/11/21/what-correlations-look-like} contains examples that clearly illustrate the extent of this phenomenon}

The correlation of metrics with human scores is highly dependent on the underlying systems used. \bleu~\cite{papineni2002bleu} has remained mostly unchanged since it was proposed in 2002, but its correlation with human scores has changed each year over ten years of evaluation (2006 to 2016) on the English--German and German--English language pairs at WMT \cite{reiter-2018-structured}. The low correlation for most of 2006--2012 is possibly due to the presence of strong rule-based systems that tend to receive low \bleu scores \cite{callison2006reevaluating}. By 2016, however, there were only a few submissions of rule-based systems, and these were mostly outperformed by statistical systems according to human judgements \cite{wmt16}. The majority of the systems in the last three years have been neural models, for which most metrics have a high correlation with human judgements. 

\bleu has been surpassed by various other metrics at every iteration of the WMT metrics shared task. Despite this, and extensive analytical evidence of the limitations of \bleu in particular and automatic metrics in general \cite{stent2005evaluating,callison2006reevaluating,smith2016climbing}, the metric remains the de facto standard of evaluating research hypotheses.  
\section{Data} 
\subsection{Direct Assessment (DA)}
Following \citet{WMT19metrics}, we use direct assessment (DA) scores  \cite{graham2017crowd} collected as part of the  human evaluation at WMT 2019. 
Annotators are asked to rate the adequacy of a set of translations compared to the corresponding source/reference sentence on a slider which maps to a continuous scale between 0 and 100. Bad quality annotations are filtered out based on quality control items included in the annotation task. Each annotator's scores are standardised to account for different scales. The score of an MT system is computed as the mean of the standardised score of all its translations.  
In WMT 19, typically around 1500--2500 annotations were collected per system for language pairs where annotator availability was not a problem. To assess whether the difference in scores between two systems is not just chance, the Wilcoxon rank-sum test is used to test for statistical significance. 
\subsection{Metrics}
Automatic metrics compute the quality of an MT output (or set of translations) by comparing it with a reference translation by a human translator. For the WMT 19 metrics task, participants were also invited to submit metrics that rely on the source instead of the reference (QE . In this paper, we focus on the following metrics that were included in evaluation at the metrics task at WMT 2019:
\subsubsection*{Baseline metrics}
\begin{compactitem} 
\item \bleu \cite{papineni-etal-2002-bleu} is the precision of $n$-grams of
  the MT output compared to the reference, weighted by a brevity penalty to
  punish overly short translations. \bleu has high variance across different hyper-parameters and pre-processing strategies, in response to which sacreBLEU \cite{post-2018-call} was introduced to create a standard implementation for all researchers to use; we use this version in our analysis. 
\item \ter \cite{snover2006ter} measures the number of edits (insertions, deletions, shifts and substitutions) required to transform the MT output to the reference. 
\item \chrf \cite{popovic2015chrf} uses character $n$-grams instead of word $n$-grams to compare the MT output with the reference. This helps with matching morphological variants of words.  
\end{compactitem}
\subsubsection*{Best metrics across language pairs}
\begin{compactitem}    
\item \yisi\cite{lo2019yisi} computes the semantic similarity of phrases in the MT output with the reference, using contextual word embeddings (BERT: \citet{devlin-etal-2019-bert}).  
\item \esim \cite{chen2016enhanced,mathur-etal-2019-putting} is a trained neural model that first computes sentence representations from BERT embeddings, then computes the similarity between the two strings.~\footnote{ESIM's submission to WMT shared task does not include scores for the language pairs en-cs and en-gu. In this paper, we use scores obtained from the same trained model that was used in the original submission.} 
\end{compactitem}
\subsubsection*{Source-based metric}
\begin{compactitem}   
\item \yisiqe \cite{lo2019yisi} is the same as \yisi, except that it uses cross-lingual embeddings to compute the similarity of the MT output with the source. 
\end{compactitem}  

The baseline metrics, particularly \bleu, were designed to use multiple references. However, in practice, they have only have been used with a single reference in recent years.


\section{Re-examining conclusions of Metrics Task 2019}

\subsection{Are metrics unreliable when evaluating high-quality MT systems?}

In general, the correlation of reference-based metrics with human scores is greater than $r = 0.8$ for all language pairs. However, the correlation is dependent on the systems that are being evaluated, and as the quality of MT increases, we want to be sure that the metrics evaluating these systems stay reliable. 

To estimate the validity of the metrics for high-quality MT systems, \citet{WMT19metrics} sorted the systems based on their Direct Assessment scores, and plotted the correlation of the top $N$ systems, with $N$ ranging from all systems to the best four systems. They found that for seven out of 18 language pairs, the correlation between metric and human scores decreases as we decrease $N$, and tends towards zero or even negative when $N = 4$.

\begin{figure*} [t]
    \centering
        (a) German--English

    \includegraphics [width=\linewidth]{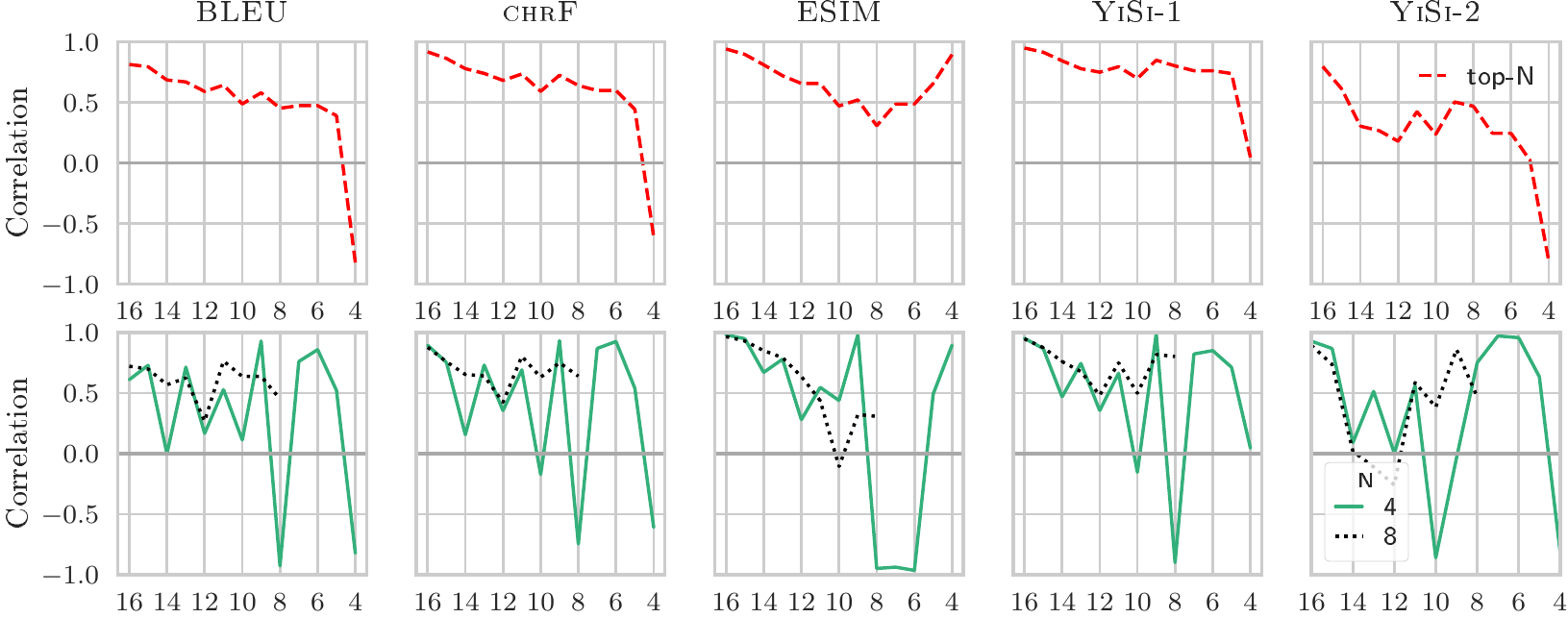} 
        (b) English--German  

    \includegraphics [width=\linewidth] {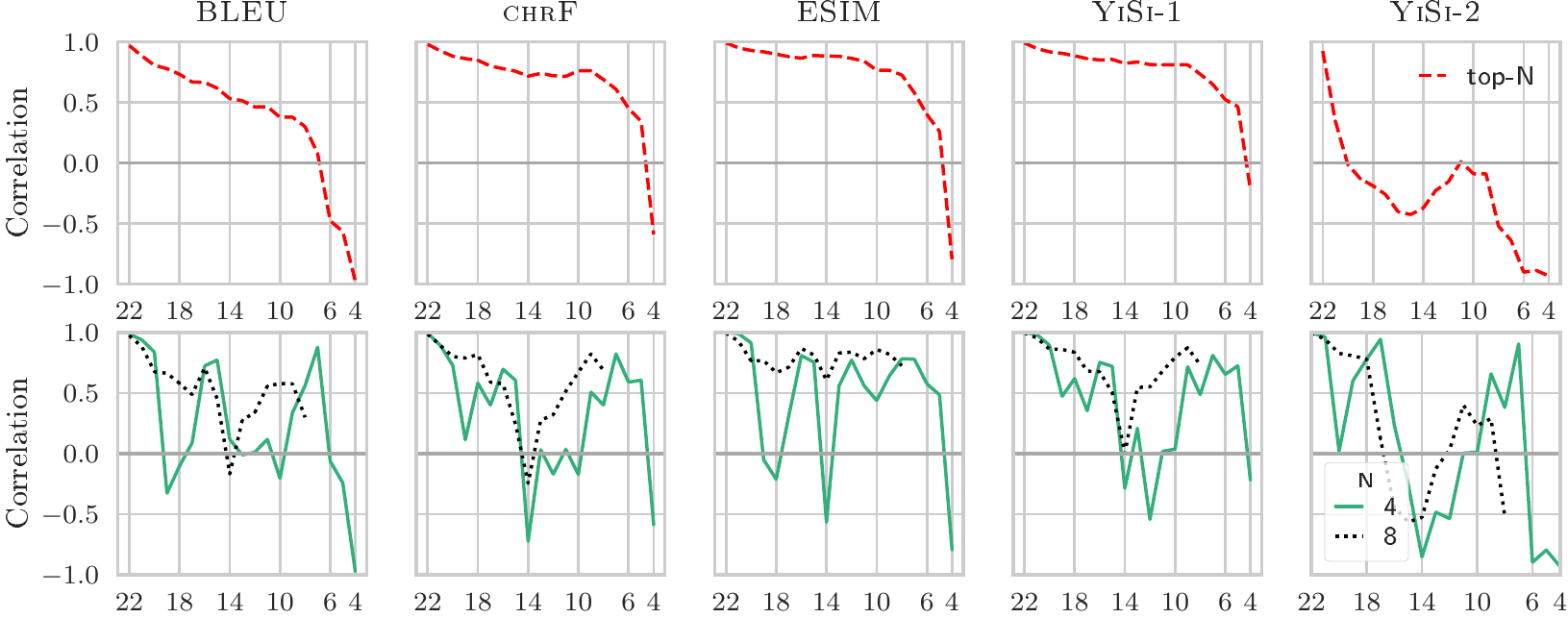} 
    \caption{Pearson correlation coefficient computed over the top-$N$ systems (top row), or over a rolling window of 4 or 8 systems (bottom row). The $x$ axis shows the index of the starting system, and systems are sorted by DA quality score.   }
    \label{fig:topn}
\end{figure*}

There are four language pairs (German--English, English--German, English--Russian, and English--Chinese) where the quality of the best MT systems is close to human performance \cite{wmt19}. If metrics are unreliable for strong MT systems, we would expect to see a sharp degradation in correlation for these language pairs. But as we look at the top $N$ systems, the correlation decreases for German--English and English--German, stays the same for English--Russian, and actually increases for English--Chinese. On the other hand, we observe this phenomenon with English--Kazakh, where the top systems are far from the quality of human translation. 

Is there another explanation for these results? Pearson's $r$ between metrics and DA scores is unstable for small samples, particularly when the systems are very close in terms of quality. The low correlation over top-$N$ systems (when $N$ is small) could  be an artefact of this instability. To understand this effect, we instead visualise the correlation of a rolling window of systems, starting with the worst $N$ systems, and moving forward by one system until we reach the top $N$ systems. The number of systems stays constant for all points in these graphs, which makes for a more valid comparison than the original setting where the sample size varies. If the metrics are indeed less reliable for strong systems, we should see the same pattern as with the top $N$ systems.

For the German--English language pair (\figref{fig:topn}b), the correlation of most metrics is very unstable when $N = 4$. 
Both \bleu and \chrf perfectly correlate with human scores for systems ranked 2--5, which then drops to $-1$ for the top 4 systems. On the other hand, \esim exhibits the opposite behaviour, even though it shows an upward trend when looking at the top-$N$ systems. 

Even worse, for English--German, \yisiqe obtains a perfect correlation at some values of $N$, when in fact its correlation with human scores is negligible once outliers are removed (\secref{sec:outliers}).

We observe similar behaviour across all language pairs: the correlation is more stable as $N$ increases, but there is no consistent trend in the correlation that depends on the quality of the systems in the sample. 
 
If we are to trust Pearson's $r$ at small sample sizes, then the reliability of metrics doesn't really depend on the quality of the MT systems. Given that the sample size is small to begin with (typically 10--15 MT systems per language pair), we believe that we do not have enough data to use this method to assess whether metric reliability decreases with the quality of MT systems.

A possible explanation for the low correlation of subsets of MT systems is that it depends on how close these systems are in terms of quality. In the extreme case, the difference between the DA scores of all the systems in the subset can be statistically insignificant, so metric correlation over these systems can be attributed to chance. 



\subsection{How do outliers affect the correlation of MT 
evaluation metrics?}
\label{sec:outliers} 

An outlier is defined as ``an observation (or subset of observations)
which appears to be inconsistent with the remainder of the dataset''
\cite{barnett1974outliers}. Pearson's $r$ is particularly sensitive to outliers in the observations. When there are systems that are generally much worse (or much better) than the rest of the systems, metrics are usually able to correctly assign low (or high) scores to these systems. In this case, the Pearson correlation can over-estimate metric reliability, irrespective of the relationship between human and metric scores of other systems.

\begin{figure} 
    \centering 
    \input{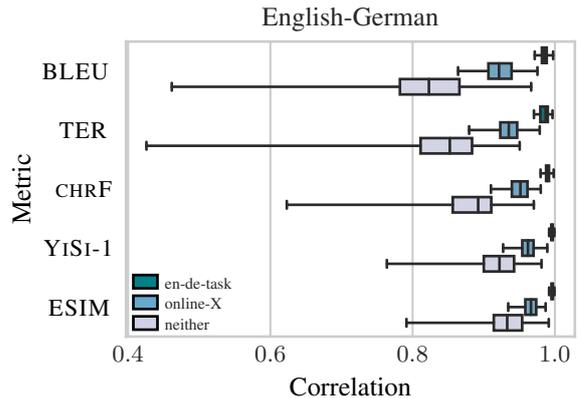}   
    \caption{Pearson's $r$ for metrics, when subsampling systems from the English--German language pair. We group the samples in the presence of the two outliers (``\texttt{en-de-task}'' and ``\texttt{Online-X}''), and when neither is present.
    }
    \label{fig:subsampl}
\end{figure}

\begin{figure*}[t!]
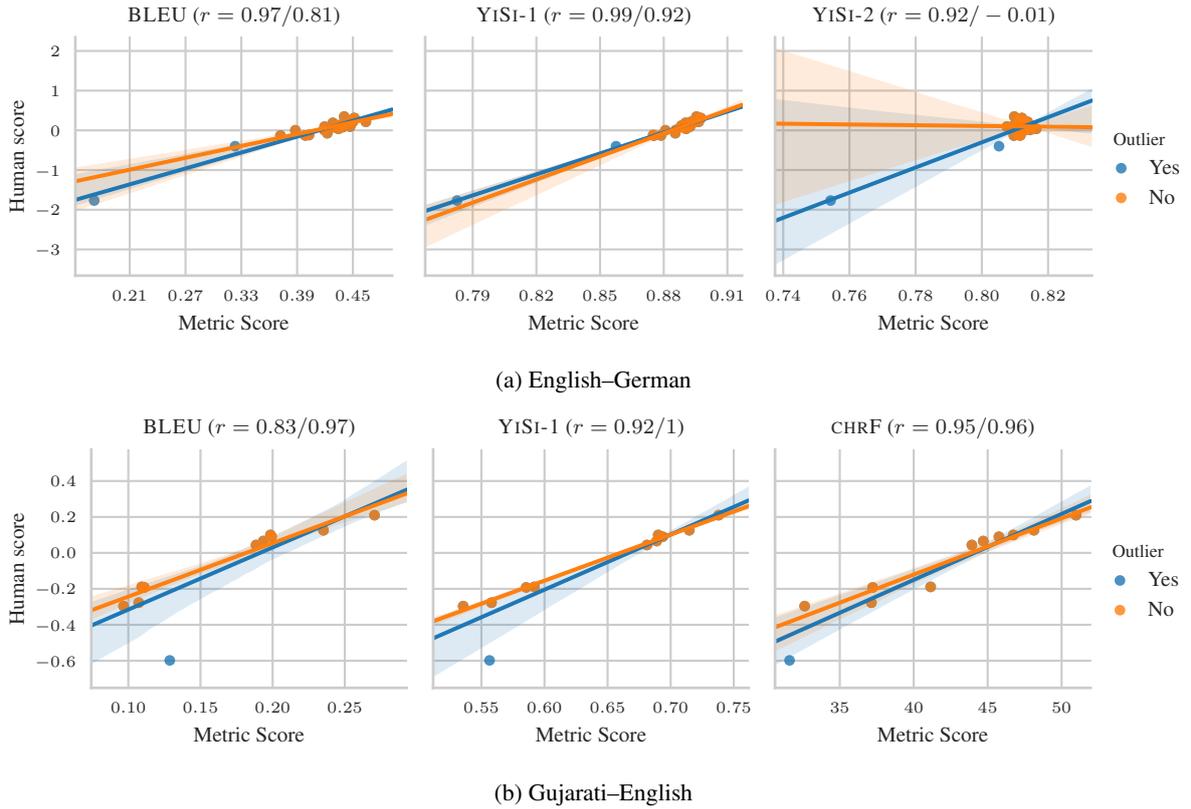

\centering
\begin{subfigure}[b]{1\textwidth}
    \input{Figs/cr-en-de-scatter.pgf}
    \caption{English--German}
    \label{fig:endescatter}
\end{subfigure}

\begin{subfigure}[b]{1\textwidth}
    \input{Figs/cr-gu-en-scatter.pgf}
    \caption{Gujarati--English}
    \label{fig:guenscatter}
\end{subfigure}

\caption{Scatter plots (and Pearson's $r$) for metrics with and without
  outliers }

\end{figure*}
 
\begin{table*}[t!]
    \centering
    \resizebox{\textwidth}{!}{

    \begin{tabular}{lrrrrrrrrrrrr}
        \toprule

 & \multicolumn{2}{c}{de--en} & \multicolumn{2}{c}{gu--en} & \multicolumn{2}{c}{kk--en} & \multicolumn{2}{c}{lt--en} & \multicolumn{2}{c}{ru--en} & \multicolumn{2}{c}{zh--en} \\
 & All & $-$out & All & $-$out & All & $-$out & All & $-$out & All & $-$out & All & $-$out \\
 \#sys & 16 & 15 & 11 & 10 & 11 & 9 & 11 & 10 & 14 & 13 & 15 & 13 \\
 \midrule
BLEU & 0.81 & 0.79 & 0.83 & 0.97 & 0.95 & 0.91 & 0.96 & 0.97 & 0.87 & 0.81 & 0.90 & 0.81  \\
TER & 0.87 & 0.81 & 0.89 & 0.95 & 0.80 & 0.57 & 0.96 & 0.98 & 0.92 & 0.90 & 0.84 & 0.72  \\ 
chrF & 0.92 & 0.86 & 0.95 & 0.96 & 0.98 & 0.77 & 0.94 & 0.93 & 0.94 & 0.88 & 0.96 & 0.84  \\
ESIM & 0.94 & 0.90 & 0.88 & 0.99 & 0.99 & 0.95 & 0.99 & 0.99 & 0.97 & 0.95 & 0.99 & 0.96  \\
YiSi-1 & 0.95 & 0.91 & 0.92 & 1.00 & 0.99 & 0.92 & 0.98 & 0.98 & 0.98 & 0.95 & 0.98 & 0.90  \\
YiSi-2 & 0.80 & 0.61 & $-$0.57 & 0.82 & $-$0.32 & 0.66 & 0.44 & 0.35 & $-$0.34 & 0.71 & 0.94 & 0.62  \\ 
\bottomrule
    \end{tabular}
    }
    \caption{Correlation of metrics with and without outliers (``All''
      and ``$-$out'', resp.) for the to-English language pairs that contain outlier systems }
    \label{tab:to_en}
\end{table*}

\begin{table*}[t!]
    \centering
     \resizebox{\textwidth}{!}{
    \begin{tabular}{lrrrrrrrrrrrr}
    \toprule
      & \multicolumn{2}{c}{de--cs} & \multicolumn{2}{c}{en--de} & \multicolumn{2}{c}{en--fi} & \multicolumn{2}{c}{en--kk} & \multicolumn{2}{c}{en--ru} & \multicolumn{2}{c}{fr--de} \\
 & All & $-$out & All & $-$out & All & $-$out & All & $-$out & All & $-$out & All & $-$out \\
 \#sys & 11 & 10 & 22 & 20 & 12 & 11 & 11 & 9 & 12 & 11 & 10 & 7 \\
 \midrule
BLEU & 0.87 & 0.74 & 0.97 & 0.81 & 0.97 & 0.94 & 0.85 & 0.58 & 0.98 & 0.95 & 0.87 & 0.85  \\ 
TER & 0.89 & 0.79 & 0.97 & 0.84 & 0.98 & 0.96 & 0.94 & 0.55 & 0.99 & 0.98 & 0.89 & 0.67  \\ 
chrF & 0.97 & 0.97 & 0.98 & 0.88 & 0.99 & 0.97 & 0.97 & 0.90 & 0.94 & 0.97 & 0.86 & 0.80  \\
ESIM & 0.98 & 0.99 & 0.99 & 0.93 & 0.96 & 0.93 & 0.98 & 0.90 & 0.99 & 0.99 & 0.94 & 0.83  \\
YiSi-1 & 0.97 & 0.98 & 0.99 & 0.92 & 0.97 & 0.94 & 0.99 & 0.89 & 0.99 & 0.98 & 0.91 & 0.85  \\
YiSi-2 & 0.61 & 0.12 & 0.92 & $-$0.01 & 0.70 & 0.48 & 0.34 & 0.69 & $-$0.77 & 0.13 & $-$0.53 & 0.07  \\
\bottomrule
    \end{tabular}
    }
    \caption{Correlation of metrics with and without outliers (``All''
      and ``$-$out'', resp.) for the language pairs into languages other than
      English that contain outlier systems.}
    \label{tab:not_to_en}
\end{table*}

Based on a visual inspection, we can see there are two outlier systems in the English--German language pair. To illustrate the influence of these systems on Pearson's $r$, we repeatedly subsample ten systems from the 22 system submissions (see \figref{fig:subsampl}). When the most extreme outlier (\texttt{en-de-task}) is present in the sample, the correlation of all metrics is greater than 0.97. The selection of systems has a higher influence on the correlation when neither outlier is present, and we can see that \yisi and \esim usually correlate much higher than \bleu.

One method of dealing with outliers is to  calculate the correlation of the rest of the points (called the skipped correlation: \citet{wilcox2004inferences}). Most of these apply methods to detect multivariate outliers in the joint distribution of the two variables: the metric and human scores in our case. However, multivariate outliers could be system pairs that indicate metric errors, and should not be removed because they provide important data about the metric.

Thus, we only look towards detecting univariate outliers based on human ratings. One common method is to simply standardise the scores, and remove systems with scores that are too high or too low. However, standardising depends on the mean and standard deviation,  which are themselves affected by outliers. Instead, we use the median and the Median Absolute Deviation (MAD) which are more robust  \cite{iglewicz1993detect,rousseeuw2011robust,leys2013detecting}.

For MT systems with human scores $s$, we use the following steps to detect outlier systems:
\begin{enumerate}
    \item Compute MAD, which is the median of all absolute
deviations from the median 
$$\text{MAD} = 1.483 \times \text{median}(|s  -  \text{median}(s)|) $$
\item compute robust scores:
 $$ z   = ( s - \text{median}(s) ) / \text{MAD} $$
\item discard systems where the magnitude of z exceeds a cutoff (we use 2.5)
\end{enumerate}

Tables \ref{tab:to_en} and \ref{tab:not_to_en} show Pearson's $r$ with and without outliers for the language pairs that contain outliers.  Some interesting observations, are as follows:

\begin{compactitem}
\item for language pairs like Lithuanian--English and English--Finnish, the correlation between
  the reference based metrics and DA is high irrespective of the presence of the outlier;
\item the correlation of BLEU with DA drops sharply from 0.85 to 0.58 for
  English--Kazakh when outliers are removed;
\item for English--German, the correlation of \bleu and \ter appears to be
  almost as high as that of \yisi and \esim. However, when we remove the
  two outliers, there is a much wider gap between the metrics.
\item if metrics wrongly assign a higher score to an outlier (e.g.\ most metrics in Gujarat--English), removing these systems increases correlation, and reporting only the skipped correlation is not ideal. 
\end{compactitem}

To illustrate the severity of the problem, we show  examples from the metrics task data where outliers present the illusion of high correlation when the metric scores are actually independent of the human scores without the outlier. For English--German, the source-based metric \yisiqe   correctly assigns a low score  to the outlier \texttt{en-de-task}. When this system is removed, the correlation is near zero. At the other extreme, \yisiqe incorrectly assigns a very high score to a low-quality outlier in the English--Russian language pair, resulting in a strongly negative  correlation. When we remove this system, we find there is no association between metric and human scores.

The results for all metrics that participated in the WMT 19 metrics task are presented in Tables 3, 4 and 5 in the appendix. 
 
\section{Beyond correlation: metric decisions for system  pairs}
\label{sec:sys_pairs}

\begin{figure}  
    \includegraphics [width=\linewidth]{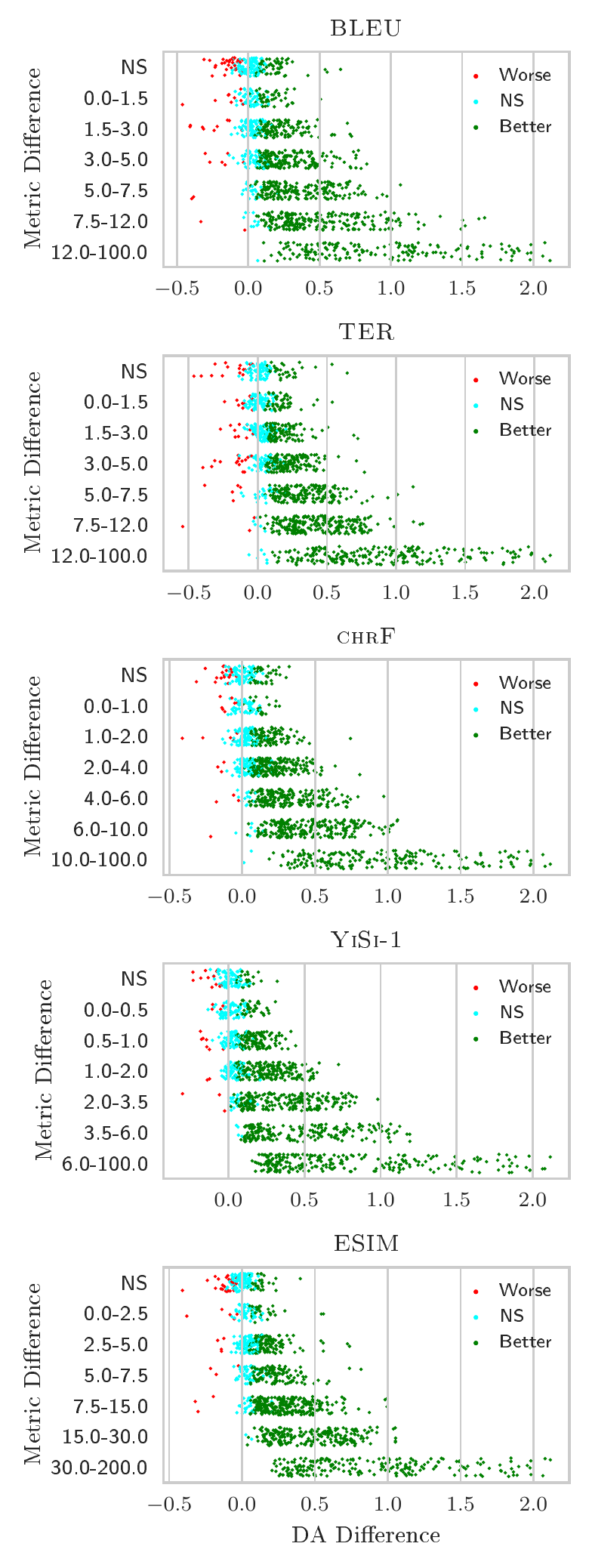} 
    \caption{Pairwise differences in human DA evaluation (x-axis) compared to difference in metric evaluation (binned on y-axis; NS means insignificant metric difference).   The colours indicate pairs judged by humans to be insignificantly different (cyan/light gray), significantly worse (red/dark gray on the left) and significantly better (green/dark gray on the right).}
    \label{fig:sp}
\end{figure} 

In practice, researchers use metric scores to compare pairs of MT systems, for instance when claiming a new state of the art, evaluating different model architectures, or even in deciding whether to publish.
Basing these judgements on metric score alone runs the risk of making wrong decisions with respect to the true gold standard of human judgements. 
That is, while a change may result in a significant improvement in \bleu, this may not be judged to be an improvement by human assessors.

Thus, we examine whether metrics agree with DA on all the MT systems pairs across all languages used in WMT 19.

Following \citet{graham2014randomized}, we use statistical significance tests to detect if the difference in scores (human or metric) between two systems (S1 and S2) can just be attributed to chance.  

For human scores, we apply the Wilcoxon rank-sum test which is used by WMT when ranking systems. We use the bootstrap method \cite{koehn-2004-statistical} to test for statistical significance of the difference in \bleu between two systems. \yisi and \esim compute the system score as the average of sentence scores, so we use the paired t-test to compute significance. Although \chrf is technically the macro-average of $n$-gram statistics over the entire test set, we treat this as a micro-average when computing   significance   such that we can use the more powerful paired t-test over sentence scores. 

Figure~\ref{fig:sp} visualises the agreement between metric score differences and differences in human DA scores.
Ideally, only differences judged as truly significant would give rise to significant and large magnitude differences under the metrics; and when metrics judge differences to be insignificant, ideally very few instances would be truly significant.
However, this is not the case: there are substantial numbers of insignificant differences even for very high metric differences (cyan, for higher range bins); moreover, the ``NS'' category --- denoting an insignificant difference in metric score --- includes many human significant pairs (red and green, top bin).
 

Considering \bleu (top plot in Figure~\ref{fig:subsampl}), for insignificant \bleu differences, humans judge one system to be better than the other for half of these system pairs. This corresponds to a Type I error. It is of concern that \bleu cannot detect these differences. Worse, the difference in human scores has a very wide range.
Conversely, when the \bleu score is significant but in the range 0--3, more than half of these systems are judged to be insignificantly different in quality (corresponding to a Type II error). For higher \bleu deltas, these errors diminish, however, even for a \bleu difference between 3 and 5 points, about a quarter of these system pairs are of similar quality.
This paints a dour picture for the utility of \bleu as a tool for gatekeeping (i.e., to define a `minimum publishable unit' in deciding paper acceptance on empirical grounds, through bounding the risk of Type II errors), as the unit would need to be whoppingly large to ensure only meaningful improvements are accepted.
Were we seek to minimise Type I errors in the interests of nurturing good ideas, the threshold would need to be so low as to be meaningless, effectively below the level required for acceptance of the bootstrap significance test.

The systems evaluated consist of a mix of systems submitted by researchers (mostly neural models) and anonymous online systems (where the MT system type is unknown). Even when we restrict the set of systems  to only neural models submitted by researchers, the patterns of Type 1 and Type 2 errors remain the same (figure omitted for space reasons).

\ter makes similar errors: \ter scores can wrongly show that a system is much better than another when humans have judged them similar, or even worse, drawn the opposite conclusion. 

\chrf, \yisi and \esim have fewer errors compared to \bleu and \ter. When these metrics mistakenly fail to detect a difference between systems, the human score difference is considerably lower than for \bleu. Accordingly, they should be used in place of \bleu. However the above argument is likely to still hold true as to their utility for gatekeeping or nurturing progress, in that the thresholds would still be particularly punitive or permissive, for the two roles, respectively. 

\begin{figure}[t!] 
    \centering 
    
    \includegraphics [width=\linewidth]{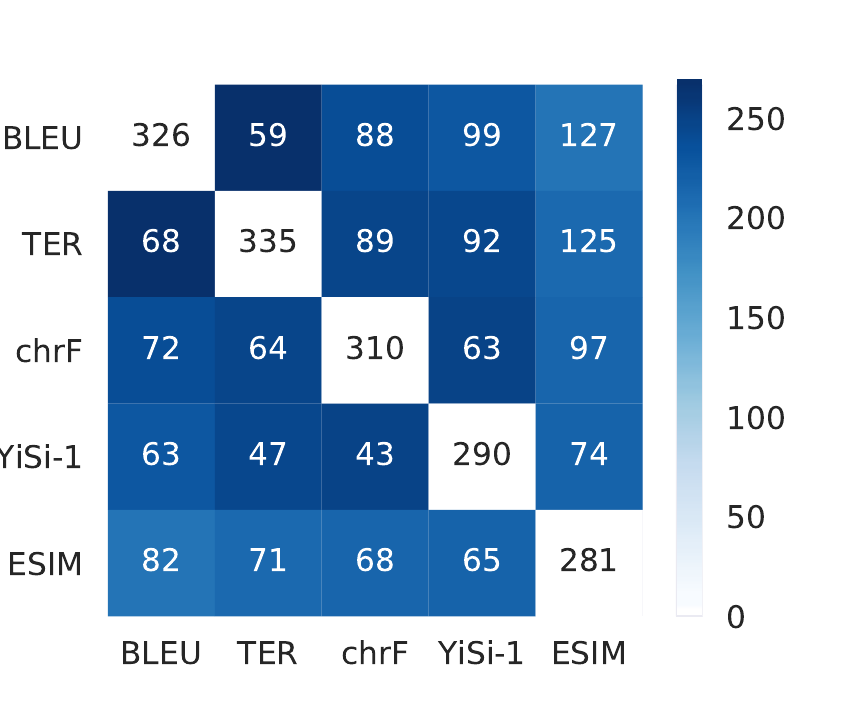}   
    \caption{The agreement between metric errors over all 1362 system comparisons. The values in the diagonal indicate the total number of Type 1 and Type 2 errors for the metric. The off-diagonal cells show the total number of errors made by the row-metric where the column-metric is correct.  }
    \label{fig:agreement}
\end{figure}

Finally, \figref{fig:agreement} looks at agreement between metric decisions when comparing MT systems. As expected, when \bleu or \ter disagree with \chrf, \esim, or \yisi, the former are more likely to be wrong. \bleu and \ter  have an 80\% overlap in errors. The decisions of \esim, a trained neural model, diverge a little more from the other metrics. Overall, despite the variety of approaches towards the task, all five metrics have common biases: over half of all erroneous decisions made by a particular metric are made in common with all other metrics. 
 

\section{Conclusion}

In this paper, we revisited the findings of the metrics task at WMT 2019, which flagged potential problems in the current best practises for assessment of evaluation metrics.

Pearson's correlation coefficient is known to be unstable for small sample sizes, particularly when the systems in consideration are very close in quality. 
This goes some way to explaining the findings whereby strong correlations between metric scores and human judgements evaporate when considering small numbers of strong systems.
We show that the same can be true for any small set of similar quality systems, not just the top systems. This effect can partly be attributed to noise due to the small sample size, rather than true shortcomings in the metrics themselves.  We need better methods to empirically test whether our metrics are less reliable when evaluating high quality MT systems.

A more serious problem, however, is outlier systems, i.e.\ those systems whose quality is much higher or lower than the rest of the systems.
We found that such systems can have a disproportionate effect on the computed correlation of metrics. 
The resulting high values of correlation can then lead to to false confidence in the reliability of metrics. Once the outliers are removed, the gap between correlation of \bleu and other metrics (e.g. \chrf, \yisi and \esim) becomes wider. In the worst case scenario, outliers introduce a high correlation when there is no association between metric and human scores for the rest of the systems. Thus, future evaluations should also measure correlations after removing outlier systems.  

Finally, the same value of correlation coefficient can describe different patterns of errors. Any single number is not adequate to describe the data, and visualising metric scores against human scores is the best way to gain insights into metric reliability. This could be done with scatter plots (e.g.\ \figref{fig:endescatter}) for each language pair, or \figref{sec:sys_pairs}, which compresses this information into one graph.
 
Metrics are commonly used to compare two systems, and accordingly we have also investigated the real meaning encoded by a difference in metric score, in terms of what this indicates about human judgements of the two systems. 
Most published work report \bleu differences of 1-2 points, however at this level we show this magnitude of difference only corresponds to true improvements in quality as judged by humans about half the time.
Although our analysis assumes the Direct Assessment human evaluation method to be a gold standard despite its shortcomings, our analysis does suggest that the current rule of thumb for publishing empirical improvements based on small \bleu differences has little meaning.

Overall, this paper adds to the case for retiring \bleu as the de facto standard metric, and instead using other metrics such as \chrf, \yisi, or \esim in its place. They are   more powerful in assessing empirical improvements. However, human evaluation must always be the gold standard, and for continuing improvement in translation, to establish significant improvements over prior work, all automatic metrics make for inadequate substitutes.

To summarise, our key recommendations are:
\begin{compactitem}
\item When evaluating metrics, use the technique outlined in \secref{sec:outliers} to remove outliers before computing Pearson's $r$.
\item When evaluating MT systems, stop using \bleu or \ter for evaluation of MT, and instead use \chrf, \yisi, or \esim;
\item Stop using small changes in evaluation metrics as the sole basis to draw important empirical conclusions, and make sure these are supported by manual evaluation.

\end{compactitem}

\section*{Acknowledgements}
We are grateful to the anonymous reviewers for their comments and valuable suggestions. This work was supported in part by the Australian Research Council.
  
 
\bibliography{anthology,acl2020}
\bibliographystyle{acl_natbib} 

\appendix
\onecolumn 
\section{The effect of removing outlier systems on the results of the WMT 19 metrics task}  
 


\begin{table*}[h!] 
    \centering
    \setlength{\tabcolsep}{2em} 
\begin{tabular}{lccc}
\toprule

& {\bf de--cs} & {\bf de--fr} & {\bf fr--de} \\
 & All\quad$-$out & All & All\quad $-$out \\
n & 11 \quad \quad10 & 11 & 10 \quad\quad 7 \\
\midrule

\metric{BEER} & {\bf 0.978}\quad0.976 & {\bf 0.941} & 0.848\quad{\bf 0.794} \\
\metric{BLEU} & {\bf 0.941}\quad0.922 & 0.891 & 0.864\quad{\bf 0.821} \\
\metric{CDER} & 0.864\quad0.734 & {\bf 0.949} & 0.852\quad{\bf 0.794} \\
\metric{CharacTER} & 0.965\quad0.959 & 0.928 & 0.849\quad{\bf 0.848} \\
\metric{chrF} & {\bf 0.974}\quad{\bf 0.970} & 0.931 & 0.864\quad{\bf 0.796} \\
\metric{chrF+} & 0.972\quad0.967 & 0.936 & 0.848\quad{\bf 0.785} \\
\metric{EED} & {\bf 0.982}\quad{\bf 0.984} & {\bf 0.940} & 0.851\quad0.792 \\
\metric{ESIM} & {\bf 0.980}\quad{\bf 0.986} & {\bf 0.950} & {\bf 0.942}\quad{\bf 0.825} \\
\metric{hLEPORa\_baseline} & 0.941\quad0.903 & 0.814 & $-$\quad\quad$-$ \\
\metric{hLEPORb\_baseline} & {\bf 0.959}\quad{\bf 0.951} & 0.814 & $-$\quad\quad$-$ \\
\metric{NIST} & {\bf 0.954}\quad0.944 & {\bf 0.916} & 0.862\quad{\bf 0.800} \\
\metric{PER} & 0.875\quad0.757 & 0.857 & {\bf 0.899}\quad0.427 \\
\metric{sacreBLE-BLEU} & 0.869\quad0.742 & 0.891 & 0.869\quad{\bf 0.846} \\
\metric{sacreBLE-chrF} & {\bf 0.975}\quad{\bf 0.980} & {\bf 0.952} & 0.882\quad{\bf 0.815} \\
\metric{TER} & 0.890\quad0.787 & {\bf 0.956} & {\bf 0.895}\quad{\bf 0.673} \\
\metric{WER} & 0.872\quad0.749 & {\bf 0.956} & {\bf 0.894}\quad{\bf 0.657} \\
\metric{YiSi-0} & {\bf 0.978}\quad0.972 & {\bf 0.952} & 0.820\quad{\bf 0.836} \\
\metric{YiSi-1} & 0.973\quad{\bf 0.980} & {\bf 0.969} & {\bf 0.908}\quad{\bf 0.846} \\
\metric{YiSi-1\_srl} & $-$\quad\quad$-$ & $-$ & {\bf 0.912}\quad0.814 \\
\hline
Source-based metrics: \\
\metric{ibm1-morpheme} & 0.355\quad0.009 & 0.509 & 0.625\quad{\bf 0.357} \\
\metric{ibm1-pos4gram} & $-$\quad\quad$-$ & 0.085 & 0.478\quad0.719 \\
\metric{YiSi-2} & 0.606\quad0.122 & 0.721 & 0.530\quad{\bf 0.066} \\
\bottomrule

\end{tabular}
\caption{Pearson correlation of metrics for the  language pairs that do not involve English. For
  language pairs that  contain outlier systems, we also show correlation after removing outlier systems. Correlations of metrics not significantly outperformed by any other for that language pair
are highlighted in bold.} 
\end{table*}


\begin{sidewaystable*}[t!]
  \smaller
    \centering
        \setlength{\tabcolsep}{1.5em}

\begin{tabular}{lccccccc}
\toprule
& {\bf de--en} & {\bf fi--en} & {\bf gu--en} & {\bf kk--en} & {\bf lt--en} & {\bf ru--en} & {\bf zh--en}   \\ 

 & All\quad$-$out & All & All\quad$-$out & All\quad$-$out & All\quad$-$out & All\quad$-$out & All\quad$-$out \\
n & 16\quad\quad15 & 12 & 11\quad\quad10 & 11\quad\quad9 & 11\quad\quad10 & 14\quad\quad13 & 15\quad\quad13 \\
\midrule
\metric{BEER} & 0.906\quad0.852 & {\bf 0.993} & 0.952\quad0.982 & 0.986\quad{\bf 0.930} & 0.947\quad0.948 & 0.915\quad0.819 & 0.942\quad0.806 \\
\metric{BERTr} & {\bf 0.926}\quad{\bf 0.897} & 0.984 & 0.938\quad{\bf 0.995} & 0.990\quad0.829 & 0.948\quad0.959 & {\bf 0.971}\quad{\bf 0.933} & 0.974\quad0.911 \\
\metric{BLEU} & 0.849\quad0.770 & 0.982 & 0.834\quad0.975 & 0.946\quad{\bf 0.912} & 0.961\quad{\bf 0.980} & 0.879\quad0.830 & 0.899\quad0.807 \\
\metric{CDER} & 0.890\quad0.827 & {\bf 0.988} & 0.876\quad0.975 & 0.967\quad0.843 & {\bf 0.975}\quad{\bf 0.981} & 0.892\quad0.875 & 0.917\quad0.847 \\
\metric{CharacTER} & 0.898\quad0.852 & {\bf 0.990} & 0.922\quad0.978 & 0.953\quad0.833 & 0.955\quad0.963 & 0.923\quad0.828 & 0.943\quad0.845 \\
\metric{chrF} & {\bf 0.917}\quad0.862 & {\bf 0.992} & 0.955\quad0.962 & 0.978\quad0.775 & 0.940\quad0.933 & 0.945\quad0.876 & 0.956\quad0.841 \\
\metric{chrF+} & {\bf 0.916}\quad0.860 & {\bf 0.992} & 0.947\quad0.961 & 0.976\quad0.769 & 0.940\quad0.934 & 0.945\quad0.878 & 0.956\quad0.851 \\
\metric{EED} & 0.903\quad0.853 & {\bf 0.994} & 0.976\quad0.988 & 0.980\quad0.779 & 0.929\quad0.930 & 0.950\quad0.872 & 0.949\quad0.840 \\
\metric{ESIM} & {\bf 0.941}\quad{\bf 0.896} & 0.971 & 0.885\quad0.986 & 0.986\quad{\bf 0.945} & {\bf 0.989}\quad{\bf 0.990} & {\bf 0.968}\quad{\bf 0.946} & {\bf 0.988}\quad{\bf 0.961} \\
\metric{hLEPORa\_baseline} & $-$ & $-$ & $-$ & 0.975\quad0.855 & $-$ & $-$ & 0.947\quad0.879 \\
\metric{hLEPORb\_baseline} & $-$ & $-$ & $-$ & 0.975\quad0.855 & 0.906\quad0.930 & $-$ & 0.947\quad0.879 \\
\metric{Meteor++\_2.0(syntax)} & 0.887\quad0.844 & {\bf 0.995} & 0.909\quad0.939 & 0.974\quad0.859 & 0.928\quad0.935 & {\bf 0.950}\quad0.878 & 0.948\quad0.836 \\
\metric{Meteor++\_2.0(syntax+copy)} & 0.896\quad0.850 & {\bf 0.995} & 0.900\quad0.930 & 0.971\quad0.871 & 0.927\quad0.931 & {\bf 0.952}\quad{\bf 0.890} & 0.952\quad0.841 \\
\metric{NIST} & 0.813\quad0.705 & 0.986 & 0.930\quad0.985 & 0.942\quad0.837 & 0.944\quad{\bf 0.963} & 0.925\quad{\bf 0.878} & 0.921\quad0.722 \\
\metric{PER} & 0.883\quad0.808 & {\bf 0.991} & 0.910\quad0.948 & 0.737\quad0.533 & 0.947\quad0.933 & 0.922\quad0.880 & 0.952\quad0.884 \\
\metric{PReP} & 0.575\quad0.452 & 0.614 & 0.773\quad0.967 & 0.776\quad{\bf 0.817} & 0.494\quad0.397 & 0.782\quad0.685 & 0.592\quad0.111 \\
\metric{sacreBLE-BLEU} & 0.813\quad0.794 & 0.985 & 0.834\quad0.975 & 0.946\quad{\bf 0.912} & 0.955\quad0.967 & 0.873\quad0.813 & 0.903\quad0.807 \\
\metric{sacreBLE-chrF} & 0.910\quad0.852 & {\bf 0.990} & 0.952\quad0.937 & 0.969\quad0.750 & 0.935\quad0.923 & 0.919\quad0.874 & 0.955\quad0.846 \\
\metric{TER} & 0.874\quad0.812 & {\bf 0.984} & 0.890\quad0.947 & 0.799\quad0.566 & 0.960\quad0.975 & 0.917\quad0.896 & 0.840\quad0.717 \\
\metric{WER} & 0.863\quad0.803 & 0.983 & 0.861\quad0.926 & 0.793\quad0.579 & 0.961\quad{\bf 0.981} & 0.911\quad0.885 & 0.820\quad0.716 \\
\metric{WMDO} & 0.872\quad{\bf 0.857} & {\bf 0.987} & 0.983\quad0.981 & {\bf 0.998}\quad{\bf 0.953} & 0.900\quad0.923 & 0.942\quad0.844 & 0.943\quad0.851 \\
\metric{YiSi-0} & 0.902\quad0.847 & {\bf 0.993} & {\bf 0.993}\quad0.990 & 0.991\quad0.876 & 0.927\quad0.933 & {\bf 0.958}\quad{\bf 0.889} & 0.937\quad0.782 \\
\metric{YiSi-1} & {\bf 0.949}\quad{\bf 0.914} & {\bf 0.989} & 0.924\quad{\bf 0.997} & 0.994\quad{\bf 0.920} & 0.981\quad0.978 & {\bf 0.979}\quad{\bf 0.947} & {\bf 0.979}\quad0.899 \\
\metric{YiSi-1\_srl} & {\bf 0.950}\quad{\bf 0.916} & {\bf 0.989} & 0.918\quad{\bf 0.998} & 0.994\quad{\bf 0.917} & {\bf 0.983}\quad{\bf 0.981} & {\bf 0.978}\quad0.943 & 0.977\quad0.897 \\
\hline
Source-based metrics: \\
\metric{ibm1-morpheme} & 0.345\quad0.223 & 0.740 & $-$ & $-$ & 0.487\quad0.638 & $-$ & $-$ \\
\metric{ibm1-pos4gram} & 0.339\quad0.137 & $-$ & $-$ & $-$ & $-$ & $-$ & $-$ \\
\metric{LASIM} & 0.247\quad0.334 & $-$ & $-$ & $-$ & $-$ & 0.310\quad0.260 & $-$ \\
\metric{LP} & 0.474\quad0.279 & $-$ & $-$ & $-$ & $-$ & 0.488\quad0.168 & $-$ \\
\metric{UNI} & 0.846\quad{\bf 0.809} & 0.930 & $-$ & $-$ & $-$ & 0.805\quad0.666 & $-$ \\
\metric{UNI+} & 0.850\quad{\bf 0.805} & 0.924 & $-$ & $-$ & $-$ & 0.808\quad0.669 & $-$ \\
\metric{YiSi-2} & 0.796\quad0.612 & 0.642 & 0.566\quad0.820 & 0.324\quad0.662 & 0.442\quad0.346 & 0.339\quad0.708 & 0.940\quad0.622 \\
\metric{YiSi-2\_srl} & 0.804\quad0.630 & $-$ & $-$ & $-$ & $-$ & $-$ & 0.947\quad0.675 \\
\bottomrule
\end{tabular}
\caption{Pearson correlation of metrics for the to-English language pairs. For
  language pairs that  contain outlier systems, we also show correlation after removing outlier systems. Correlations of metrics not significantly outperformed by any other for that language pair
are highlighted in bold.}
\end{sidewaystable*}

\begin{sidewaystable*}[t!]
  \smaller
    \centering
            \setlength{\tabcolsep}{1.5em}

  \begin{tabular}{lcccccccc}
\toprule
& {\bf en--cs} & {\bf en--de} & {\bf en--fi} & {\bf en--gu} & {\bf en--kk} & {\bf en--lt} & {\bf en--ru} & {\bf en--zh} \\
 & All & All\quad$-$out & All\quad$-$out & All & All\quad$-$out & All & All\quad$-$out & All \\
n & 11 & 22\quad\quad20 & 12\quad\quad11 & 11 & 11\quad\quad9 & 12 & 12\quad\quad11 & 12 \\
\midrule
\metric{BEER} & {\bf 0.990} & 0.983\quad0.869 & {\bf 0.989}\quad{\bf 0.978} & 0.829 & 0.971\quad0.826 & {\bf 0.982} & 0.977\quad0.947 & 0.803 \\
\metric{BLEU} & 0.897 & 0.921\quad0.419 & {\bf 0.969}\quad0.943 & 0.737 & 0.852\quad0.576 & {\bf 0.989} & 0.986\quad0.967 & 0.901 \\
\metric{CDER} & 0.985 & 0.973\quad0.849 & {\bf 0.978}\quad{\bf 0.957} & 0.840 & 0.927\quad0.668 & {\bf 0.985} & {\bf 0.993}\quad{\bf 0.981} & 0.905 \\
\metric{CharacTER} & {\bf 0.994} & {\bf 0.986}\quad{\bf 0.886} & 0.968\quad0.939 & {\bf 0.910} & 0.936\quad{\bf 0.895} & 0.954 & {\bf 0.985}\quad{\bf 0.982} & 0.862 \\
\metric{chrF} & 0.990 & 0.979\quad0.881 & {\bf 0.986}\quad{\bf 0.972} & {\bf 0.841} & {\bf 0.972}\quad{\bf 0.900} & {\bf 0.981} & 0.943\quad0.968 & 0.880 \\
\metric{chrF+} & {\bf 0.991} & 0.981\quad0.883 & {\bf 0.986}\quad{\bf 0.970} & {\bf 0.848} & {\bf 0.974}\quad{\bf 0.907} & {\bf 0.982} & 0.950\quad0.973 & 0.879 \\
\metric{EED} & {\bf 0.993} & {\bf 0.985}\quad{\bf 0.894} & {\bf 0.987}\quad{\bf 0.978} & {\bf 0.897} & {\bf 0.979}\quad{\bf 0.883} & 0.975 & 0.967\quad{\bf 0.984} & 0.856 \\
\metric{ESIM} & $-$ & {\bf 0.991}\quad{\bf 0.928} & 0.957\quad0.926 & $-$ & {\bf 0.980}\quad{\bf 0.900} & {\bf 0.989} & {\bf 0.989}\quad{\bf 0.986} & 0.931 \\
\metric{hLEPORa\_baseline} & $-$ & $-$ & $-$ & 0.841 & 0.968\quad{\bf 0.852} & $-$ & $-$ & $-$ \\
\metric{hLEPORb\_baseline} & $-$ & $-$ & $-$ & 0.841 & 0.968\quad0.852 & 0.980 & $-$ & $-$ \\
\metric{NIST} & 0.896 & 0.321\quad0.246 & 0.971\quad0.936 & 0.786 & 0.930\quad0.611 & {\bf 0.993} & {\bf 0.988}\quad{\bf 0.973} & 0.884 \\
\metric{PER} & 0.976 & 0.970\quad0.815 & {\bf 0.982}\quad{\bf 0.961} & 0.839 & 0.921\quad0.545 & 0.985 & 0.981\quad0.955 & 0.895 \\
\metric{sacreBLE-BLEU} & {\bf 0.994} & 0.969\quad0.806 & {\bf 0.966}\quad0.939 & 0.736 & 0.852\quad0.576 & {\bf 0.986} & 0.977\quad0.946 & 0.801 \\
\metric{sacreBLE-chrF} & 0.983 & 0.976\quad0.874 & 0.980\quad0.958 & 0.841 & {\bf 0.967}\quad0.840 & 0.966 & {\bf 0.985}\quad{\bf 0.988} & 0.796 \\
\metric{TER} & 0.980 & 0.969\quad0.841 & {\bf 0.981}\quad{\bf 0.960} & {\bf 0.865} & 0.940\quad0.547 & {\bf 0.994} & {\bf 0.995}\quad{\bf 0.985} & 0.856 \\
\metric{WER} & 0.982 & 0.966\quad0.831 & {\bf 0.980}\quad{\bf 0.958} & {\bf 0.861} & 0.939\quad0.525 & {\bf 0.991} & {\bf 0.994}\quad{\bf 0.983} & 0.875 \\
\metric{YiSi-0} & {\bf 0.992} & 0.985\quad0.869 & {\bf 0.987}\quad{\bf 0.977} & 0.863 & 0.974\quad0.840 & 0.974 & 0.953\quad{\bf 0.967} & 0.861 \\
\metric{YiSi-1} & 0.962 & {\bf 0.991}\quad{\bf 0.917} & 0.971\quad0.937 & {\bf 0.909} & {\bf 0.985}\quad{\bf 0.892} & 0.963 & {\bf 0.992}\quad{\bf 0.978} & {\bf 0.951} \\
\metric{YiSi-1\_srl} & $-$ & {\bf 0.991}\quad{\bf 0.917} & $-$ & $-$ & $-$ & $-$ & $-$ & {\bf 0.948} \\
\hline
Source-based metrics: \\
\metric{ibm1-morpheme} & 0.871 & 0.870\quad0.198 & 0.084\quad0.254 & $-$ & $-$ & 0.810 & $-$ & $-$ \\
\metric{ibm1-pos4gram} & $-$ & 0.393\quad0.449 & $-$ & $-$ & $-$ & $-$ & $-$ & $-$ \\
\metric{LASIM} & $-$ & 0.871\quad0.007 & $-$ & $-$ & $-$ & $-$ & 0.823\quad0.336 & $-$ \\
\metric{LP} & $-$ & 0.569\quad0.558 & $-$ & $-$ & $-$ & $-$ & 0.661\quad0.178 & $-$ \\
\metric{UNI} & 0.028 & 0.841\quad0.251 & 0.907\quad0.808 & $-$ & $-$ & $-$ & 0.919\quad0.760 & $-$ \\
\metric{UNI+} & $-$ & $-$ & $-$ & $-$ & $-$ & $-$ & 0.918\quad0.746 & $-$ \\
\metric{USFD} & $-$ & 0.224\quad0.301 & $-$ & $-$ & $-$ & $-$ & 0.857\quad0.514 & $-$ \\
\metric{USFD-TL} & $-$ & 0.091\quad0.212 & $-$ & $-$ & $-$ & $-$ & 0.771\quad0.177 & $-$ \\
\metric{YiSi-2} & 0.324 & 0.924\quad0.014 & 0.696\quad0.478 & 0.314 & 0.339\quad{\bf 0.685} & 0.055 & 0.766\quad0.134 & 0.097 \\
\metric{YiSi-2\_srl} & $-$ & 0.936\quad0.155 & $-$ & $-$ & $-$ & $-$ & $-$ & 0.118 \\
\bottomrule
\end{tabular}
\caption{Correlation of metrics for the from-English language pairs. For
  language pairs that  contain outlier systems, we also show correlation after removing outlier systems. Values in bold indicate that the metric is not
  significantly outperformed by any other metric under the Williams Test.}
\end{sidewaystable*}

\end{document}